\documentclass[a4paper,fleqn]{cas-dc}

\usepackage[numbers]{natbib}

\usepackage{amsmath}
\usepackage{subcaption}
\usepackage{caption}
\usepackage{tabularx, array}
\usepackage{fontawesome}
\usepackage[dvipsnames]{xcolor}
\usepackage{tikz}
\usepackage{seqsplit}

\definecolor{checkgreen}{RGB}{80,142,40}
\definecolor{crossred}{RGB}{184,0,0}

\newcommand*\Circle[1]{\protect\tikz[baseline=(char.base)]{
            \protect\node[shape=circle,draw,inner sep=0.5pt] (char) {\footnotesize{#1}};}}

\newcommand{\checkicon}{\textcolor{checkgreen}{\faCheckCircle}}
\newcommand{\crossicon}{\textcolor{crossred}{\faTimesCircle}}
\begin{document}
\let\WriteBookmarks\relax
\def\floatpagepagefraction{1}
\def\textpagefraction{.001}

\shorttitle{}    

\shortauthors{}  

\title [mode = title]{Geometric deep learning for local growth prediction on abdominal aortic aneurysm surfaces}  




\author[1]{Dieuwertje Alblas}[orcid=0000-0002-7754-7405]
\cormark[1]
\cortext[1]{Corresponding author}
\ead{d.alblas@utwente.nl}
\author[1]{Patryk Rygiel}
\author[1]{Julian Suk}
\author[2,4]{Kaj O. Kappe}
\author[1]{Marieke Hofman}
\author[1]{Christoph Brune}
\author[2,3,4]{Kak Khee Yeung}
\author[1]{Jelmer M. Wolterink}



\affiliation[1]{organization={Department of Applied Mathematics, Technical Medical Centre, University of Twente},
            addressline={Drienerlolaan 5}, 
            city={Enschede},
%          citysep={}, % Uncomment if no comma needed between city and postcode
            postcode={7522 NB},
            country={The Netherlands}}

\affiliation[2]{organization={Department of Surgery, Amsterdam University medical center, Location University of Amsterdam,},
            addressline={Meibergdreef 9}, 
            city={Amsterdam},
%          citysep={}, % Uncomment if no comma needed between city and postcode
            postcode={1105 AZ}, 
            country={The Netherlands}}

\affiliation[3]{organization={Department of Surgery, Amsterdam University medical center, location Vrije Universiteit Amsterdam,},
            addressline={Boelelaan 1117}, 
            city={Amsterdam},
%          citysep={}, % Uncomment if no comma needed between city and postcode
            postcode={1081 HV}, 
            country={The Netherlands}}

\affiliation[4]{organization={Amsterdam Cardiovascular Sciences, Atherosclerosis and Aortic Diseases}, 
            city={Amsterdam},
%          citysep={}, % Uncomment if no comma needed between city and postcode
            country={The Netherlands}}


\begin{abstract}
Abdominal aortic aneurysms (AAAs) are progressive focal dilatations of the abdominal aorta. AAAs may rupture, with a survival rate of only 20\%.  Current clinical guidelines recommend elective surgical repair when the maximum AAA diameter exceeds 55 mm in men or 50 mm in women. Patients that do not meet these criteria are periodically monitored, with surveillance intervals based on the maximum AAA diameter. However, this diameter does not take into account the complex relation between the 3D AAA shape and its growth, making standardized intervals potentially unfit. Personalized AAA growth predictions could improve monitoring strategies. We propose to use an SE(3)-symmetric transformer model to predict AAA growth directly on the vascular model surface enriched with local, multi-physical features. In contrast to other works which have parameterized the AAA shape, this representation preserves the vascular surface's anatomical structure and geometric fidelity. We train our model using a longitudinal dataset of 113 computed tomography angiography (CTA) scans of 24 AAA patients at irregularly sampled intervals. After training, our model predicts AAA growth to the next scan moment with a median diameter error of 1.18 mm. We further demonstrate our model's utility to identify whether a patient will become eligible for elective repair within two years (acc = 0.93). Finally, we evaluate our model's generalization on an external validation set consisting of 25 CTAs from 7 AAA patients from a different hospital. Our results show that local directional AAA growth prediction from the vascular surface is feasible and may contribute to personalized surveillance strategies.
\end{abstract}

%\nocite{*}

% Keywords
\maketitle

% Main text
\section{Introduction}\label{sec:intro}
% Wat zijn AAAs en hoe wordt daar nu mee omgegaan in de kliniek?
Abdominal aortic aneurysms (AAAs) are focal dilatations of the abdominal aorta that exceed 30 mm in diameter. If left untreated, AAAs may rupture with fatal consequences in >80\% of cases~\citep{reimerink2013systematic}. To prevent rupture, elective repair is recommended when the maximum AAA diameter exceeds 55 mm in men or 50 mm in women, or when the growth rate exceeds 10 mm/year \citep{wanhainen2019editor,wanhainen2024editor}. AAA diameters are typically measured via non-invasive imaging, such as duplex ultrasound (DUS) or computed tomography angiography (CTA).

Patients with AAAs below the repair threshold undergo routine surveillance, with follow-up intervals depending on the AAA diameter. Current guidelines recommend monitoring every three years in AAAs with diameters of 30-39 mm, decreasing to six months in AAAs with diameters of 50-55 mm \citep{wanhainen2024editor}. While these guidelines are motivated by increased growth rates in larger AAAs, diameter alone is an imperfect predictor of individual growth rates, which could lead to inadequate surveillance schedules \citep{powell2011systematic} and patient anxiety \citep{lee2017experience}. Personalized prediction of AAA growth could help customize surveillance intervals and provide AAA patients with peace of mind.

Aneurysm growth is highly individualized and is believed to be governed by a complex, non-linear process influenced by many patient-specific features, e.g., sex, age, smoking status and diabetes \citep{brady2004abdominal}, the presence and thickness of an intraluminal thrombus (ILT) \citep{martufi2016local,zhu2020intraluminal}, hemodynamic quantities such as wall shear stress (WSS) \citep{stevens2017biomechanical} and inflammation of the vessel wall~\citep{ma3rs2017aortic} or surrounding tissue \citep{meekel2021inflammatory}. A personalized AAA growth prediction model should be able to incorporate such characteristics.

\begin{figure*}[t!]
    \centering
    \includegraphics[width=\textwidth]{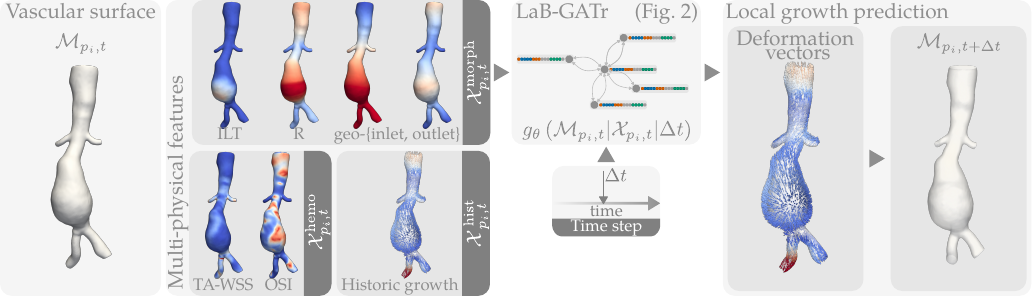}
    \caption{Our local AAA growth prediction method. We use a transformer $g_{\theta}$ that predicts local AAA growth over a time step $\Delta t$ given $\mathcal{M}_{p_i, t}$, a mesh model of the AAA vasculature with multi-physical features $\mathcal{X}_{p_i, t}$ embedded on the vertices. For each vertex, $g_{\theta}$ estimates a 3D deformation vector. Together, these vectors prescribe the deformation of $\mathcal{M}_{p_i, t}$ into $\mathcal{M}_{p_i, t+\Delta t}$.}
    \label{fig:method}
\end{figure*}

Previous studies have used machine learning (ML) to relate patient-specific features and AAA growth. A common approach is to predict the future maximum diameter using regression or classification models based on 1D patient feature vectors. These features could include current diameter \citep{ristl2022growth}, geometric features \citep{lindquist2021geometric,chandrashekar2023prediction} or hemodynamic parameters \citep{lee2018applied,forneris2023ai}. However, representing the patient as a 1D feature vector and AAA growth as a single quantity ignores the spatial heterogeneity of AAA growth \citep{braet2021vascular} and the effect of localized features on AAA growth.

To account for spatial heterogeneity, local AAA growth can be predicted by convolutional neural networks (CNNs) after projecting the vascular geometry and features to a Euclidean grid. One approach is to parameterize the AAA along its centerline and represent growth as changes in the 1D diameter profile \citep{jiang2020deep}. Another approach is to project the vascular surface and local features to a 2D plane, parameterized by the centerline and azimuthal angle. Given this 2D plane and features, AAA growth can be predicted as local changes in radius \citep{kim2022deep}. While these methods allow for a more detailed representation of AAA growth, they still rely on flattening the vascular geometry into a Euclidean grid, which could introduce distortions and limit the flexibility of the representation.

Alternatively, geometric deep learning models have been proven effective in predicting vector fields directly on the 3D vascular model, obviating the need to project the vascular model to a Euclidean grid \citep{suk2024physics,suk2024mesh,do2018prediction}. These geometric deep learning models are able to process signals that live on non-Euclidean structures, such as vascular surfaces. However, the orientation and location of these vascular surfaces in 3D space is arbitrary. As AAA growth is mostly driven by intrinsic biomechanical and hemodynamic forces rather than external forces like gravity, the predicted growth should not be affected by the orientation of the aorta in a scan. The model can gain some robustness to variations in orientation and position of the vascular surface in 3D space through data augmentation. Alternatively, robustness of the deep learning model is guaranteed when it is SE(3)-symmetric. This property ensures that the model predictions rotate along with the orientation of the vascular surface and are unaffected by translations of the vascular surface. Moreover, SE(3)-symmetry significantly improves the model's data efficiency \citep{suk2023se}.

AAA growth is a continuous process and to model this accurately, a growth prediction model should be able to predict growth at any future time point. This flexibility is also needed for analyzing and learning from real-world data, as patients are typically scanned at highly irregular intervals. We propose temporal augmentation to obtain a model that is robust to variations of these intervals. Furthermore, a predictive model that can be evaluated enables denser temporal sampling of the growth trajectory. This enables clinicians to anticipate when the AAA may reach a critical size and can hence be used to support personalized surveillance intervals.

In this work, we develop a method for predicting AAA growth over an arbitrary time interval directly on the 3D vascular surface. Our contributions are:
\begin{enumerate}
    \item We propose a geometric deep learning model to predict local AAA growth patterns directly on the vascular surface.
    \item We embed multi-physical, patient specific features that have been associated with AAA growth into the vascular surface.
    \item We condition our growth prediction model on arbitrary time intervals, allowing personalized growth predictions at any time within a prediction horizon.
    \item We introduce temporal data augmentation with neural fields \citep{alblas2023implicit, garzia2024neural} to allow training on a real-life dataset of AAAs measured at irregular intervals.
\end{enumerate}

\section{Methods \& Materials}
Figure \ref{fig:method} visualizes our problem statement and proposed solution. We consider the surface of the abdominal aorta, including the AAA, and its surrounding vasculature as a manifold that evolves continuously in time. We represent AAA growth as a time-dependent 3D vector field acting on the manifold that describes the pointwise deformation. Consequently, we represent patient $p_i$ at time point $t$ by a 3D model of the outer wall of their abdominal aorta, iliac- and renal arteries, denoted by the manifold $\mathcal{M}_{p_i,t}$. On this manifold, we can construct $d$-dimensional signals $\mathcal{X}_{p_i,t}(v): \mathcal{M}_{p_i,t} \to \mathbb{R}^d$. The signals we consider represent multi-physical features that can be derived from image data and affect AAA growth. We include patient-specific morphology, hemodynamics and historical growth, which we denote by $\mathcal{X}_{p_i,t}^{\text{morph}}(v)$, $\mathcal{X}_{p_i,t}^{\text{hemo}}(v)$ and $\mathcal{X}_{p_i,t}^{\text{hist}}(v)$, respectively. Note that these three functions represent vector fields defined on the full surface of the vascular model. We concatenate these vectors vertex-wise and denote this as $\mathcal{X}_{p_i,t}(v)$. Given the vascular model $\mathcal{M}_{p_i,t}$, $\mathcal{X}_{p_i,t}$, and a time step $\Delta t$, we aim to find the deformation of $\mathcal{M}_{p_i,t}$ into $\mathcal{M}_{p_i, t+\Delta t}$. In other words, we wish to find the operator $\mathcal{G}(\mathcal{M}_{p_i,t} | \mathcal{X}_{p_i,t} | \Delta t):\mathcal{M}_{p_i, t} \mapsto \mathcal{M}_{p_i, t+\Delta t}$.

We approximate the operator $\mathcal{G}$ by a neural network $g_\theta$. Without loss of generality, we the discretize vascular manifold $\mathcal{M}_{p_i,t}$ into a triangular mesh consisting of $n_{p_i,t}$ vertices. On each vertex $v \in \mathcal{M}_{p_i,t}$, we embed the local multi-physical features $\mathcal{X}^{\text{morph}}_{p_i,t}(v)$, $\mathcal{X}^{\text{hemo}}_{p_i,t}(v)$, and $\mathcal{X}^{\text{hist}}_{p_i,t}(v)$. Moreover, we describe the geometry of $\mathcal{M}_{p_i,t}$ in every vertex as the local tangent plane. As shown in Figure \ref{fig:method}, the network $g_{\theta}$ is conditioned on these multi-physical features, as well as the geometry of $\mathcal{M}_{p_i,t}$ and $\Delta t$.
For each vertex, $g_{\theta}$ estimates a 3D deformation vector; together, these vectors prescribe the deformation of the vascular model $\mathcal{M}_{p_i,t}$ into $\mathcal{M}_{p_i,t + \Delta t}$. Our growth prediction model is trained using a real-world dataset of longitudinal vascular models, acquired at irregularly spaced intervals in time.

\subsection{Growth prediction model}
\begin{figure*}[t!]
\centering
    \includegraphics[scale=1]{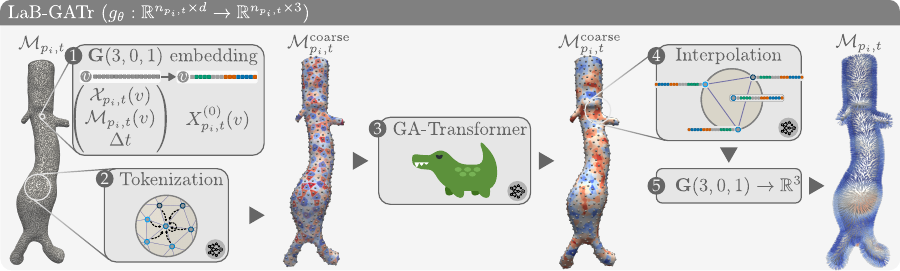}
    \caption{Overview of the LaB-GATr architecture $g_{\theta}$, adapted from \citep{suk2024lab}. Patient-specific features, local geometry and time step are embedded per vertex as multi-vectors in $\mathbf{G}(3,0,1)$ (\Circle{1}). Next, $\mathcal{M}_{p_i,t}$ is tokenized into $\mathcal{M}^{\text{coarse}}_{p_i,t}$ (\Circle{2}). Tokens are fed into the geometric algebra transformer (GATr) (\Circle{3}), whose output is interpolated to the resolution of $\mathcal{M}_{p_i,t}$ (\Circle{4}). Lastly, the network outputs are projected back to $\mathbb{R}^3$ (\Circle{5}).
    }
    \label{fig:labgatr}
\end{figure*}

As neural network $g_{\theta}$, we use the LaB-GATr architecture introduced in \citep{suk2024lab}, operating directly on the vertices $v \in \mathcal{M}_{p_i,t}$ and their embedded multi-physical features. Figure \ref{fig:labgatr} shows an overview of the LaB-GATr architecture, consisting of three trainable parts. First, there is a tokenization module (\Circle{2}) that downsamples the mesh model and creates tokens at this coarser scale. Second, the geometric algebra transformer (GATr) (\Circle{3}) as introduced in \citep{brehmer2024geometric} that processes these tokens. Third, the interpolation module (\Circle{4}) that upsamples the GATr output to the original mesh resolution. The up- and downsampling steps are essential for efficient processing of large meshes. LaB-GATr operates on mesh vertices with features embedded in the projective geometric algebra $\mathbf{G}(3,0,1)$. Before serving as input to the network, features are lifted to $\mathbf{G}(3,0,1)$ (\Circle{1}) and after processing, network outputs are projected back to the original, Euclidean domain (\Circle{5}). 

\subsubsection{Geometric algebra embedding}
LaB-GATr operates on input features embedded in the projective geometric algebra $\mathbf{G}(3,0,1)$. A unique property of this space is the joint representation of geometric objects such as scalars, vectors and planes and geometric operations such as translations and rotations as 16-dimensional \textit{multi-vectors}:
\begin{equation}\label{eq:multivector}
\begin{aligned}
    x \hspace{1pt} = \hspace{1pt} (x_s, \underbrace{x_0, x_1, x_2, x_3}_{\text{vectors}}, \underbrace{x_{01}, x_{02}, x_{03}, x_{12}, x_{13}, x_{23}}_{\text{bivectors}}, \\
    \underbrace{x_{012}, x_{013}, x_{023}, x_{123}}_{\text{trivectors}}, x_{0123})
\end{aligned}
\end{equation}
Each of these geometric objects has a predefined embedding in $\mathbf{G}(3,0,1)$, as shown in Table \ref{tab:GA_embedding}. We use these embeddings for our vascular models $\mathcal{M}_{p_i,t}$, $\mathcal{X}_{p_i,t}$ and $\Delta t$ in $\mathbf{G}(3,0,1)$ (\Circle{1} in Figure \ref{fig:labgatr}). For each vertex $v \in \mathcal{M}_{p_i,t}$, we embed and concatenate $c$ local features to construct a $d = c \cdot 16$ dimensional feature vector $X^{(0)}(v)$. Specifically, we use $c=9$, using 7 channels to embed the multi-physical features; 4, 2 and 1 channels to embed $\mathcal{X}^{\text{morph}}_{p_i, t}$, $\mathcal{X}^{\text{hemo}}_{p_i, t}$ and $\mathcal{X}^{\text{hist}}_{p_i, t}$ (Figure \ref{fig:method}) as scalars (Table \ref{tab:GA_embedding}), respectively. The remaining 2 channels are used to embed  $\Delta t$ as a scalar and the geometry of $\mathcal{M}_{p_i,t}$ as the tangent plane in vertex $v$ as an oriented plane (Table \ref{tab:GA_embedding}). We denote the $\mathbf{G}(3,0,1)$ embedding of the full vascular mesh model by the tensor $X^{(0)}_{p_i,t} \in \mathbb{R}^{n_{p_i} \times d}$.

\begin{table}[h!]
    \centering
    \caption{Embedding of scalars, planes and points and translations as 16-dimensional multi-vectors in $\mathbf{G}(3,0,1)$ (Eq. \eqref{eq:multivector}), the remaining elements of $x$ remain zero. Embedding is performed in \Circle{1} (Fig. \ref{fig:labgatr}).}
    \label{tab:GA_embedding}
    \begin{tabularx}{0.5\textwidth}{@{} >{\arraybackslash}m{0.5\columnwidth} >{\arraybackslash}m{0.5\columnwidth} }
    \hline
    Geometric object / operation & $\mathbf{G}(3,0,1)$ embedding \\
    \hline
    Scalar $s \in \mathbb{R}$ & $x_s = s$ \\
    Plane with normal $\nu \in \mathbb{R}^3$, offset $\delta\in\mathbb{R}$ & $(x_0, x_1, x_2, x_3) = (\delta, \nu)$ \\
    Point $\rho \in \mathbb{R}^{3}$ & $(x_{012}, x_{013}, x_{023}, x_{123}) = (\rho, 1)$ \\
    \hline
    Translation $\tau \in \mathbb{R}^3$ & $(x_s, x_{01}, x_{02}, x_{03}) = (1, \frac{1}{2} \tau)$\\
    \hline
    \end{tabularx}
\end{table}

  \subsubsection{Transformer}\label{sec:transformer}
 At the core of the LaB-GATr architecture is a transformer model (\Circle{3} in Figure \ref{fig:labgatr}), that consists of geometric algebra transformer (GATr) blocks that respect the structure of $\mathbf{G}(3,0,1)$ \citep{brehmer2024geometric}. The $l^{\text{th}}$ transformer block updates a given input $X^{(l)}$ as follows:
 \begin{equation*}
    \begin{aligned}
     A^{(l)} &= X^{(l)} + \\
     &  \hspace{-2pt} \xi \left(\underset{h}{\text{Concat}} \hspace{3pt} \text{Softmax} \left( \frac{q_h(X^{(l)}) k_h(X^{(l)})^T}{\sqrt{d/2}}\right) v_h (X^{(l)})\right) \\
     X^{(l+1)} &= A^{(l)} + \text{MLP}(A^{(l)}).
     \end{aligned}
 \end{equation*}
We use multi-head self-attention to concatenate outputs over multiple heads $h$ \citep{vaswani2017attention}, followed by a learned linear map $\xi$. The query, key and value maps $q_h$, $k_h$ and $v_h$ ($\mathbb{R}^{n\times d} \rightarrow \mathbb{R}^{n\times d}$) consist of layer normalization composed with learned linear maps. The feature maps that come out of these transformer blocks are equivariant under rotations and translations encoded in $X^{(l)}$. This implies that for any group action $\rho \in SE(3)$, the layer maps $\rho X^{(l)} \mapsto \rho X^{(l+1)}$.

\subsubsection{Tokenization and interpolation}\label{sec:tokens}
% Pooling + interpolation 
In order to efficiently process the full vascular mesh models, LaB-GATr performs pooled tokenization across local clusters of mesh vertices (\Circle{2} in Figure \ref{fig:labgatr}), prior to passing it to the GA-Transformer (\Circle{3} in Figure \ref{fig:labgatr}, Section \ref{sec:transformer}). To obtain these tokens, we first perform farthest point subsampling of the vertices of $\mathcal{M}_{p_i,t}$ to obtain a coarser mesh, i.e. $\mathcal{M}^{\text{coarse}}_{p_i,t}$, with $m_{p_i,t} = \eta \cdot n_{p_i,t}$ vertices. Next, we cluster the $v \in \mathcal{M}_{p_i,t}$ based on the closest vertex $p \in \mathcal{M}^{\text{coarse}}_{p_i,t}$:
\begin{align*}
    C(p) = \{ v \in \mathcal{M}_{p_i,t} | p = \underset{q \in \mathcal{M}_{p_i,t}^{\text{coarse}}}{\text{argmin}} \Vert q - v \Vert \}.
\end{align*}
Tokens $X^{(1)}(p)$ are obtained per vertex $p \in \mathcal{M}^{\text{coarse}}_{p_i,t}$ by learned message passing of $X^{(0)}(v)$ from $v \in C(p)$:
\begin{equation*}
    \begin{aligned}
    m_{v \to p} &= \text{MLP}(X^{(0)}(v), p-v)\\
    X^{(1)}(p) &= \frac{1}{|C(p)|} \sum\limits_{v\in C(p)} m_{v\to p},
    \end{aligned}
\end{equation*}
where the MLP is a multi-layer perceptron as introduced in \citep{brehmer2024geometric}. The feature vectors of $p \in \mathcal{M}^{\text{coarse}}_{p_i,t}$ together form the tensor $X^{(1)} \in \mathbb{R}^{m_{p_i,t} \times d}$ that serves as input for the transformer (\Circle{3} in Figure \ref{fig:labgatr}).

% interpolation
The transformer (\Circle{3}) outputs a prediction at the resolution of $\mathcal{M}^{\text{coarse}}_{p_i,t}$. In order to retrieve the resolution of $\mathcal{M}_{p_i,t}$, we perform learned, proportional interpolation to obtain the output $Y(v)$ at $v \in \mathcal{M}_{p_i,t}$ (\Circle{4} in Figure \ref{fig:labgatr}):
\begin{equation*}
    \begin{aligned}
        X^{(l+1)}(v) &= \frac{\sum_p \lambda_{p,v} X^{(l)}(p)}{\sum_p \lambda_{p,v}}, \hspace{5pt} \lambda_{p,v}:=\frac{1}{\Vert p-v \Vert} \\
        Y(v) &= \text{MLP}(X^{(l+1)}(v), X^{(0)}).
    \end{aligned}
\end{equation*}
For each $v \in \mathcal{M}_{p_i,t}$ we sum over the three closest $p\in \mathcal{M}^{\text{coarse}}_{p_i,t}$ and obtain the output for all $n_{p_i,t}$ vertices in $\mathcal{M}_{p_i,t}$, i.e. $Y \in \mathbb{R}^{n_{p_i,t}\times 3}$.

\begin{figure}[t!]
    \centering
    \includegraphics[width=0.5\textwidth]{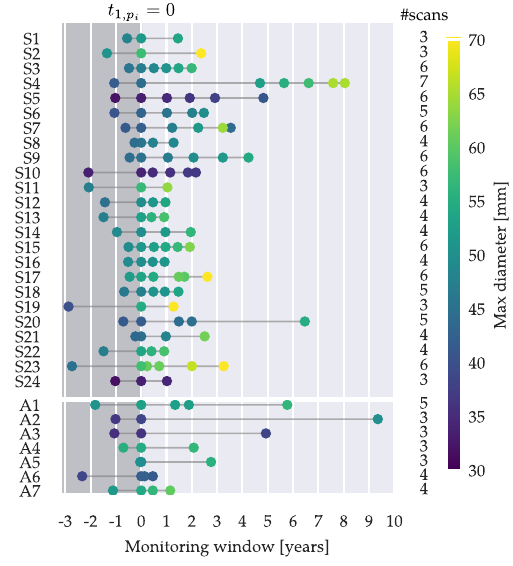}
    \caption{Surveillance intervals, number of scans and maximum diameters of the AAAs for patients in our training dataset (Seoul National University hospital, South Korea) \citep{jiang2020deep} and our external validation set (Amsterdam UMC, The Netherlands), shown in the top and bottom plots, respectively.}
    \label{fig:data_overview}
\end{figure}

\subsection{Data}
% irregulier gespacede data in de tijd.
Our vascular models $\mathcal{M}_{p_i, t}$ and patient-specific features $\mathcal{X}_{p_i,t}$ are acquired from a set of retrospectively included pre-operative 3D computed tomography angiographies (CTAs). For each patient, the dataset contains at least three scans, taken at $m_{p_i}$ irregularly spaced intervals. We denote the set of follow-up time points for patient $p_i$ by $T_{p_i} = \{t_{0,p_i}, t_{1,p_i}, ..., t_{m_{p_i} - 1, p_i} \}$, with $t_{1,p_i}=0$, see Figure \ref{fig:data_overview}. To simplify notation, we omit the patient indexing in individual time points, e.g. $t_1, t_2$ instead of $t_{1, p_i}, t_{2, p_i}$. 

The longitudinal AAA data used in this work consists of 138 CTAs collected from 31 AAA patients from two different sources. Our training dataset consists of 113 CTAs from 24 AAA patients which were obtained from Seoul National University Hospital, South Korea \citep{jiang2020deep,kim2022deep,do2018prediction}. The remaining 25 CTA's from 7 AAA patients were obtained from the Amsterdam UMC hospital in Amsterdam, The Netherlands and serve as an independent test set. Figure \ref{fig:data_overview} shows variation in the surveillance intervals and maximum diameters between patients S1-S24 and A1-A7. The median time step size between two successive scans is 0.91 (0.58) years and 1.09 (1.69) for patients S1-S24 and A1-A7, respectively. The median annual diameter increments are respectively 2.86 (3.22) mm and 1.09 (2.27) mm.

In each CTA, we segment standardized regions of the lumen of both common iliac and renal arteries and the lumen and ILT of the abdominal aorta \citep{rygiel2024global, isensee2021nnu}. From these segmentations, we obtain mesh models $\mathcal{M}_{p_i, t}$ of the artery outer walls, consisting of 65,000 vertices on average. As AAA growth is a slow and subtle process, it is a prerequisite that the mesh models for each patient are well aligned. Therefore, we rigidly align the meshes by the centerlines of their abdominal aortas using the iterative closest point algorithm.\\

\subsection{Continuous AAA growth}\label{sec:inr}
The intervals between acquired scans in our dataset are irregular. In order to use arbitrary intervals during training, normalize time-dependent features, and perform temporal augmentation, we interpolate growth from discrete and irregularly spaced time points $t \in T_{p_i}$ \citep{garzia2024neural}.

\begin{figure}[t!]
    \centering
    \includegraphics[width=0.5\textwidth]{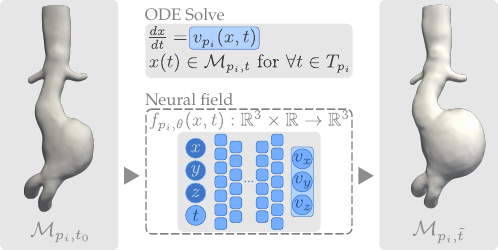}
    \caption{Continuous representation of AAA growth for an individual patient. We approximate the velocity field $v_{p_i}(x,t)$ by a neural field $f_{p_i,\theta}$ conditioned on spatio-temporal coordinates \textit{(bottom)}. We solve a neural ODE \textit{(top)} involving $f_{p_i,\theta}$ to deform $\mathcal{M}_{p_i,0}$ into the vascular model at any $\tilde{t}\in [t_{0}, t_{m_{p_i} - 1} ]$.}
    \label{fig:neural-ODE}
\end{figure}

We construct a continuous representation of AAA growth for each patient in our dataset. We model the deformation of the original vascular model $\mathcal{M}_{p_i,t_0}$ as a diffeomorphic transformation, described by a time-dependent velocity field $v_{p_i}(x,t)$ \citep{beg2005computing}. The deformation follows an ordinary differential equation (ODE): 
\begin{equation*}
\frac{dx}{dt} = v_{p_i}(x,t), \text{ for } x\in\mathcal{M}_{p_i,t_0}
\end{equation*}
We approximate the velocity field $v_{p_i,t}(x,t)$ by a neural field $f_{p_i, \theta}(x,t)$, that is conditioned on continuous spatio-temporal coordinates \citep{sitzmann2020implicit}. $f_{p_i, \theta}$ is optimized using the vascular manifold at time points $t \in T_{p_i}$. As shown in Figure \ref{fig:neural-ODE}, $f_{p_i, \theta}$ outputs a velocity field for each queried spatio-temporal coordinate. The vascular model at any time point $\tilde{t} \in [t_{0}, t_{m_{p_i} - 1}]$ can be obtained by solving the following initial value problem:
\begin{align}\label{eq:neural-ode}
    \frac{dx}{dt} = f_{p_i,\theta}(x,t), \hspace{5pt} x(0) \in \mathcal{M}_{p_i, t_0}.
\end{align}

To optimize the neural field $f_{p_i,\theta}$, we solve the initial value problem in Equation \eqref{eq:neural-ode} as a neural ODE \citep{chen2018neural}, using all vascular models of patient $p_i$ at time points $t \in T_{p_i}$. During training, we deform the vertices $v$ at location $x \in \mathcal{M}_{p_i, t_0}$ to their positions at later time points by solving ODE \eqref{eq:neural-ode} \citep{garzia2024neural}. We use the Chamfer distance between the estimated and ground-truth vertex positions at $\mathcal{M}_{p_i,t}, t\in T_{p_i}$ as a loss function to train $f_{p_i, \theta}$:
\begin{equation}\label{eq:chamfer}
\begin{aligned}
    \mathcal{L}_{\text{Chamfer}}\left(P, Q \right) =
    \frac{1}{|P|} \sum\limits_{p \in P} \min\limits_{q \in Q} \Vert p - q \Vert \\ 
     + \frac{1}{|Q|} \sum\limits_{q \in Q} \min\limits_{p \in P} \Vert p - q \Vert,
\end{aligned}
\end{equation}
where $P$ are the deformed vertices of $\mathcal{M}_{p_i,t_0}$ and $Q$ are the vertices of $\mathcal{M}_{p_i,t}, t\in T_{p_i}$. Note that $\mathcal{M}_{p_i,t_0}$ and $\mathcal{M}_{p_i,t}$ have a different numbers of vertices. After training, the model $f_{p_i, \theta}$ can be used to obtain the vascular surface $\mathcal{M}_{p_i, \tilde{t}}$ at any time point $\tilde{t} \in [t_0, t_{m_{p_i} - 1}]$.\\

\subsection{Multi-physical features}
We consider multi-physical features that locally describe \textit{morphology}, \textit{hemodynamics} and \textit{historical growth} on the vascular surface models. We obtain these features from the vascular models $\mathcal{M}_{p_i,t}$ and their estimated continuous deformations $f_{p_i, \theta}$.

\subsubsection{Morphological features}
The morphological features $\mathcal{X}^{\text{morph}}_{p_i,t}$ consist of three different parameters, as shown in Figure \ref{fig:method}. First, we obtain the ILT thickness, which has shown positive correlations to AAA growth \citep{parr2011thrombus}. We determine the shortest distance between the outer wall and the lumen for each vertex $v \in \mathcal{M}_{p_i,t}$. Second, we determine the local radius of the vascular model using the inscribed sphere method as a function of the vascular model centerlines \citep{gharahi2015growth}. We project these radii to $v \in \mathcal{M}_{p_i,t}$ by identifying the closest centerline point \citep{antiga2008image}. Third, we use the geodesic distances to the inlets and outlets of the vascular model as features, that we determine using the heat method \citep{crane2017heat} for each vertex in $\mathcal{M}_{p_i,t}$. Although these geodesics are not directly correlated to AAA growth, they have been successfully used as geometric descriptors in previous studies \citep{suk2024mesh,suk2024lab}. These geodesics intrinsically describe the position of the vascular model's inlet and outlet, giving our SE(3)-symmetric model a sense of orientation which is otherwise lost due to its symmetry preserving properties.

\subsubsection{Hemodynamic features}
Hemodynamic features have been hypothesized to be associated with AAA growth \citep{groeneveld2018systematic}. We include the time-averaged wall shear stress (TAWSS) and oscillatory shear index (OSI) in $\mathcal{X}^{\text{hemo}}$. Low TAWSS correlates with AAA growth \citep{bappoo2021low}, but the effect of high TAWSS remains ambiguous \citep{staarmann2019shear}. High OSI values have been associated with an increased risk for aortic expansion \citep{song2023systematic}.

To obtain both features, we perform computational fluid dynamics (CFD) simulations with pulsatile blood flow in the lumen of each vascular model using SimVascular \citep{updegrove2017simvascular}. We assume rigid walls with no-slip boundary conditions and impose a standardized pulsatile, parabolic boundary condition at the inlet \citep{bazilevs2006isogeometric}. We run the simulations for three cardiac cycles on a CPU cluster (32 cores, 4-6 hours per simulation) and obtain patient-specific velocity and pressure fields. TAWSS is computed by integrating the wall shear stress (WSS) magnitude over a cardiac cycle, while OSI quantifies directional changes in the WSS. Both hemodynamic features are projected from the lumen to the outer wall surface $\mathcal{M}_{p_i,t}$ using the closest point on the lumen surface for each vertex $v \in \mathcal{M}_{p_i,t}$.\\

\subsubsection{Historic growth features}
Including historic growth patterns have significantly contributed to growth prediction performance in previous work \cite{kim2022deep}. We compute the historic growth feature $\mathcal{X}^{\text{hist}}_{p_i,t}$ (Figure \ref{fig:method}) from the personalized, continuous AAA growth models $f_{p_i, \theta}$. This scalar feature locally describes the magnitude of AAA growth over the past six months. For each $t \geq 0 \in T_{p_i}$, we approximate the vascular model six months earlier using $f_{p_i, \theta}$. $\mathcal{X}^{\text{hist}}_{p_i,t}$ is obtained for each $v \in \mathcal{M}_{p_i,t}$ as the distance to the closest vertex on $\mathcal{M}_{p_i, t-6 \text{ months}}$.

\begin{figure}[t!]
    \centering
    \includegraphics[width=0.5\textwidth]{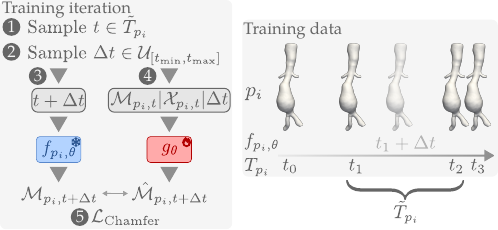}
    \caption{Training procedure for $g_{\theta}$ using temporal augmentation. First, we randomly sample a time point $t$ from $\tilde{T}_{p_i}$ (\Circle{1}) and a time step $\Delta t \in \mathcal{U}_{\left[ t_{\text{min}}, t_{\text{max}}\right]}$ (\Circle{2}). Next, we acquire the target vascular model $\mathcal{M}_{p_i, t + \Delta t}$ using $f_{p_i,\theta}$ (\Circle{3}). Then, we let $g_{\theta}$ predict the growth from $\mathcal{M}_{p_i,t}$, $\mathcal{X}_{p_i,t}$ and $\Delta t$ to obtain $\hat{\mathcal{M}}_{p_i,t + \Delta t}$ (\Circle{4}). The weights of $g_{\theta}$ are updated using the Chamfer loss between $\mathcal{M}_{p_i,t + \Delta t}$ and $\hat{\mathcal{M}}_{p_i,t + \Delta t}$ (\Circle{5})}
    \label{fig:model training}
\end{figure}

\subsection{Model training} \label{sec:optimization}
Our growth prediction model $g_{\theta}$ predicts the deformation of a vascular model $\mathcal{M}_{p_i, t}$, given the features $\mathcal{X}_{p_i, t}$ and an arbitrary time step $\Delta t$. Since the available patient data consists of irregularly spaced time points, learning to predict growth at arbitrary intervals is challenging. To address this, we train our model with \textit{temporal augmentation} using the continuous AAA growth representations $f_{p_i,\theta}$ (Section \ref{sec:inr}).

As illustrated in Figure \ref{fig:model training}, training $g_\theta$ with temporal augmentation consists of five steps. First, we randomly sample a time point $t \in \tilde{T}_{p_i} := \{ t_1,..., t_{m_{p_i} - 2}\}$ for the input vascular model $\mathcal{M}_{p_i, t}$ (\Circle{1}). Note that this can never be an interpolated shape, as we do not have the associated multi-physical features $\mathcal{X}_{p_i,t}$ at these time points. Second, we sample a time step $\Delta t \sim \mathcal{U}_{\left[ t_{\text{min}}, t_{\text{max}}\right]}$ (\Circle{2}), where hyperparameters $t_{\text{min}}$ and $t_{\text{max}}$ define the range of time steps used during training. Third, we acquire the target vascular model $\mathcal{M}_{p_i, t + \Delta t}$ from $f_{p_i, \theta}$ (\Circle{3}). Fourth, we let $g_{\theta}$ predict the growth to obtain vascular model $\hat{\mathcal{M}}_{p_i, t+\Delta t}$ (\Circle{4}). Fifth, we compute the Chamfer distance (Eq. \eqref{eq:chamfer}) between $\hat{\mathcal{M}}_{p_i, t+\Delta t}$ and $\mathcal{M}_{p_i, t+\Delta t}$ and use this to update the weights of $g_{\theta}$, while the weights of $f_{p_i, \theta}$ are frozen (\Circle{5}).

\subsection{Growth prediction baselines}\label{sec:baselines}
We introduce two baselines to benchmark the performance of our growth prediction model $g_{\theta}$. The first baseline assumes no growth, i.e. zero deformation vectors in each vertex of the vascular model \citep{chen2024vestibular}. As AAA growth is a very slow process, predicting no growth is a reasonable baseline for short intervals, that we denote by $B^{\text{zero}}$. Second, we use the historical growth from continuous model $f_{p_i, \theta}$ to \textit{extrapolate} local growth patterns. We use the deformation from the past six months proportional to $\Delta t$ as the vertex-wise deformation vectors. This method, which we denote by $B^{\text{hist}}$, can be considered as a linear extrapolation of the growth.

\begin{figure}[t!]
    \centering
    \includegraphics[width=0.5\textwidth]{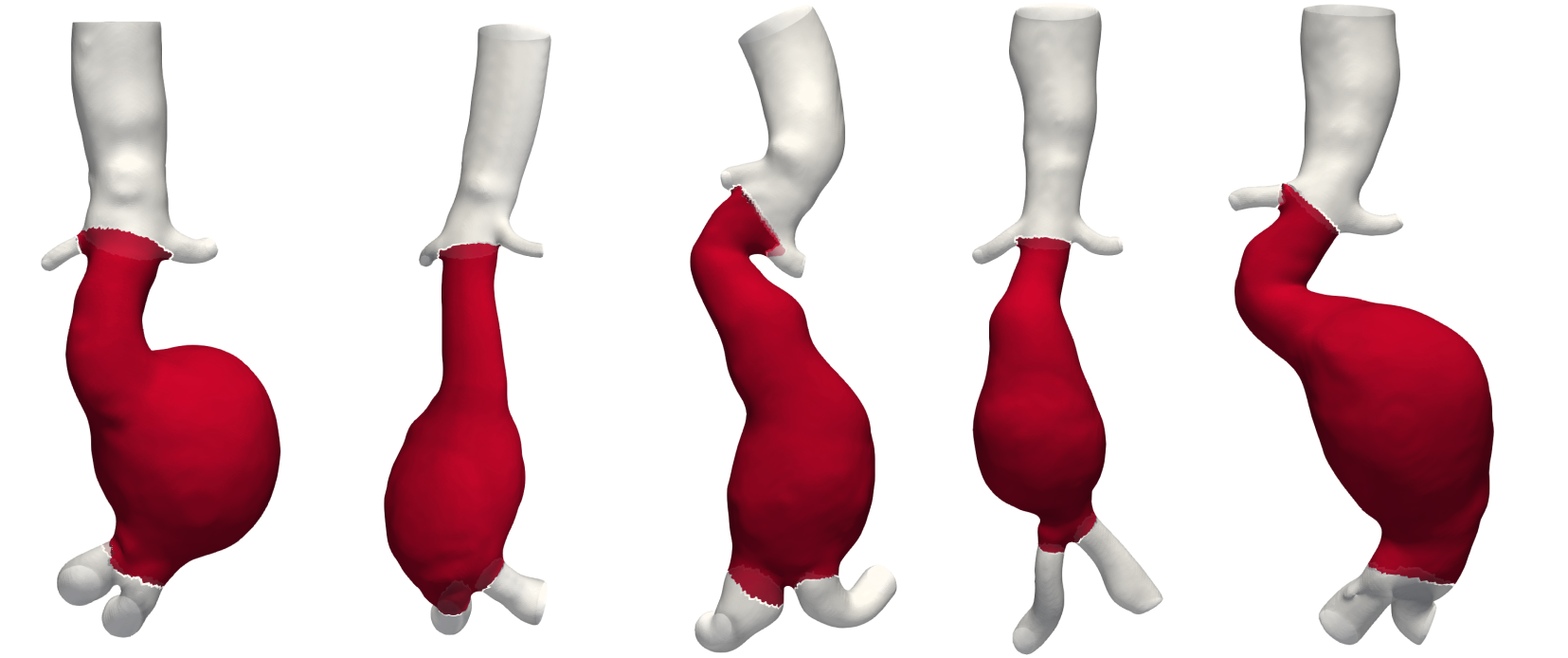}
    \caption{Vascular models of five patients in our training dataset, with the infrarenal aorta, the section of the vascular model used to evaluate the performance of our growth prediction model, marked in red.}
    \label{fig:infrarenal}
\end{figure}

\subsection{Performance metrics}\label{sec:evaluation}
We evaluate the performance of our growth prediction model both qualitatively and quantitatively. For all evaluations, unless explicitly stated otherwise, we use only the vascular models acquired from patient scans at time points $t \in T_{p_i}$ to evaluate the predicted growth i.e., we do not compare against interpolated vascular models. We quantitatively assess the growth prediction on the infrarenal part of the aorta, shown in Figure \ref{fig:infrarenal}.

We calculate multiple complementary metrics to capture different aspects of aneurysm progression:
\begin{itemize}
    \item \textbf{Local surface distances:} We compute the local accuracy of the AAA growth predictions by computing the local surface distances between each vertex on the predicted and reference vascular model. We use these distances for a visual assessment of the growth prediction. Additionally, we compute the 95\textsuperscript{th} percentile Hausdorff distance (HD95).
    \item \textbf{Growth volume:} To quantify overall aneurysm expansion, we compute the volume of the infrarenal section at each time point. We define the growth volume as the difference in these volumes between consecutive time points, which can be negative if the AAA shrinks.
    \item \textbf{Relative growth volume difference}: Since AAA growth is gradual, absolute growth volumes are small and difficult to compare between patients. To account for this, we define the relative growth volume difference (RGVD) as the difference between the predicted and reference growth volumes, normalized by the refrence growth volume. Underestimation of the actual growth results in a negative RGVD.
    \item \textbf{Diameters:} Maximum AAA diameters remain the clinical gold standard for measuring aneurysm progression \citep{wanhainen2019editor}. We extract maximum diameters from the vascular models orthogonal to the centerline and use diameter profiles along the centerline for qualitative evaluation. We define the diameter error as the absolute difference between the predicted and target maximum diameter.
\end{itemize}

\begin{figure*}
    \centering
    \includegraphics[width=\textwidth]{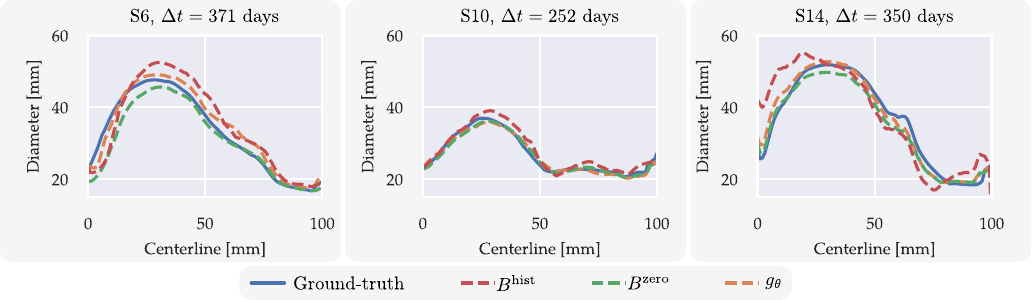}
    \caption{Diameter profiles of three different patients from the cross-validation cohort (Fig. \ref{fig:data_overview}), predicted with our model $g_{\theta}$ and baseline methods $B^{\text{zero}}$, $B^{\text{hist}}$.}
    \label{fig:diameters_baselines}
\end{figure*}

\section{Experiments \& Results}
For our continuous AAA growth representations $f_{p_i, \theta}$, we use an MLP with periodic activation functions \citep{sitzmann2020implicit}, which are trained using an Adam optimizer for 500 epochs using a learning rate of 3e-3. We train the growth prediction model $g_{\theta}$ on patients S1-S24 (Figure \ref{fig:data_overview}) using 5-fold cross-validation, that we split at patient-level ($N = 24$). We downsample the vascular mesh models with a rate of $\eta = 0.05$ (Section \ref{sec:tokens}). Our model is trained with temporal augmentation (Section \ref{sec:optimization}) using randomly sampled time steps $\Delta t$ between $t_{\text{min}} = 6$ months and $t_{{\text{max}}} = 2$ years (\Circle{2} in Fig. \ref{fig:model training}). We use the Chamfer distance on the abdominal aorta as the loss function and an Adam optimizer with a learning rate of 3e-4 and a batch size of four for 2,000 epochs to train $g_{\theta}$ on an NVIDIA L40 GPU (48GB). 

\subsection{Growth prediction performance}\label{sec:exp_1}
We evaluate our growth prediction model $g_{\theta}$ on the cross-validation patient cohort and compare it to the two baseline methods $B^{\text{zero}}$ and $B^{\text{hist}}$ (Sec \ref{sec:baselines}). For all 24 patients, we assess the growth predictions between successive time points $t_{i, p_j}$ and $t_{i+1,j}$ for $i=1,2,...,m_{p_j} - 2$. Note that this excludes growth predictions between $t_0$ and $t_1$, as historical growth features are not available at $t_0$. This results in 65 vascular models used for performance assessment.

\begin{table}[t!]
\caption{Growth prediction performance in terms of the 95\textsuperscript{th} percentile Hausdorff distance, diameter error and relative growth volume difference using the two baselines $B^{\text{zero}}$ and $B^{\text{hist}}$ and our model $g_{\theta}$. Median and IQR values are shown for each metric.}
\label{tab:baselines}
\setlength{\tabcolsep}{2pt} 
\begin{tabular}{ccccccc}
\hline
\multirow{2}{*}{Method} & \multicolumn{2}{c}{HD95 {[}mm{]}} & \multicolumn{2}{c}{Diameter error {[}mm{]}} & \multicolumn{2}{c}{RGVD} \\
 & median & IQR & median & IQR & median & IQR \\
 \hline
$B^{\text{zero}}$ & 2.97 & 2.12 & 2.50 & 3.22 & -1.00 & 0.00 \\
$B^{\text{hist}}$ & 3.97 & 3.49 & 2.28 & 3.70 & 0.46 & 1.47 \\
$g_{\theta}$ & \textbf{2.75} & 1.96 & \textbf{1.18} & 1.38 & \textbf{0.10} & 0.92\\
\hline
\end{tabular}%
\end{table}

For all growth predictions, we compute the 95\textsuperscript{th} percentile Hausdorff distance, the diameter error and the relative growth value difference. Table \ref{tab:baselines} shows that our model $g_{\theta}$ outperforms both baselines on all three metrics. The relative growth value difference clearly differs for each method: as expected, $B^{\text{zero}}$ consistently underestimates growth, with a constant RGVD of -1.00. Conversely, $B^{\text{hist}}$ tends to overestimate growth with a median RGVD of 0.46, whereas our model achieves a lower median RGVD of 0.10. Moreover, our model outperforms the baseline models on diameter error and HD95. Figure \ref{fig:diameters_baselines} shows diameter profiles for growth predictions of three patients in the dataset. These diameter plots show that the local AAA growth patterns predicted by $g_{\theta}$ are also closer to the actual growth than both baselines. Moreover, these plots confirm that $B^{\text{hist}}$ tends to overestimate growth.

\begin{figure}[t!]
    \centering
    \includegraphics[width=0.5\textwidth]{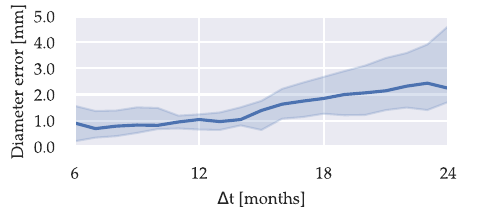}
    \caption{Median and IQR diameter error of the growth prediction as a function of the time step $\Delta t$.}
    \label{fig:diameter_density}
\end{figure}

\subsection{Time conditioning}
Our growth prediction model is conditioned on an arbitrary time step $\Delta t$, varying between six months and two years during training. We assess the effect of the time step on the growth prediction performance during inference. Starting from the second vascular model, i.e. $\mathcal{M}_{p_i, t_1}$, we let $g_{\theta}$ predict the vascular model for $\Delta t \in \{6, 7, 8,...,24\}$ months. We computed the maximum diameter for the predicted vascular model and compare this with the target vascular model obtained from $f_{p_i,\theta}$. Note that in this experiment we evaluate performance based on interpolated shapes, as the vascular models are not available from the ground-truth data at these exact time points.

Figure \ref{fig:diameter_density} shows the median diameter errors and their IQRs as a function of $\Delta t$. We observe that the diameter error increases as the time step $\Delta t$ increases, but stays below 2.5 mm for $\Delta t = 24$ months.

\begin{figure}
    \centering
    \includegraphics[width=0.5\textwidth]{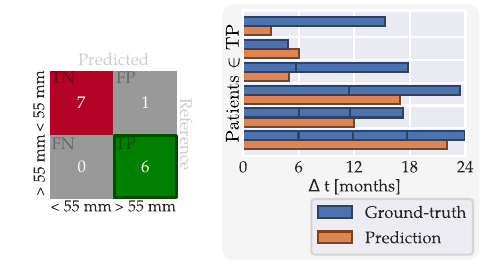}
    \caption{Prediction of whether and when the AAA diameter exceeds 55 mm within a two-year period. \textit{Left}: Confusion matrix showing the performance of $g_{\theta}$ in classifying whether the diameter threshold is exceeded within two years. \textit{Right}: Comparison of predicted and observed time intervals until the AAA exceeds 55 mm, for the 6 patients where this occurred. For observed data, scan moments are indicated by vertical lines.}
    \label{fig:intervals_timeline}
\end{figure}

Accurate prediction of whether and when a AAA will exceed the 55 mm threshold is essential for tailoring patient-specific surveillance intervals. To evaluate this, we assess if our model can predict if the AAA diameter exceeds this threshold within a two year time-frame and estimate the corresponding time window. We use patients from the cross-validation cohort with a AAA diameter below 55 mm at $t_1$ and either exceed the threshold within two years or have at least two years of remaining follow-up data ($N=14$). The median diameter at $t_1$ was 44.7 (8.7) mm. Based on the observed scan data, 6 patients exceeded the 55 mm threshold within two years, while 8 did not.

We let our growth prediction model estimate the deformation over the next two years with one month increments and checked if the 55 mm threshold is exceeded. As shown in Figure \ref{fig:intervals_timeline} (left), our model correctly identified all 6 patients whose aneurysms exceeded the threshold and 7 of the 8 who did not, leading to an overall accuracy of 0.93.

For the 6 positive cases, we compare the predicted time at which the AAA diameter exceeds 55 mm to the timing observed from the scan data, as shown in Figure \ref{fig:intervals_timeline} (right). Because the clinical scan intervals are relatively long, the exact moment when the AAA crosses the threshold is not observed directly but must lie between the penultimate and final scan. Conversely, the predicted AAA diameters are evaluated at a monthly frequency, allowing for finer estimation of this event. Consistent with this, the predicted time until the AAA exceeds 55 mm is shorter than the observed time until this event for most patients. For one patient, the predicted time slightly exceeded the observed interval.

\begin{figure}[t!]
    \centering
    \includegraphics[width=0.5\textwidth]{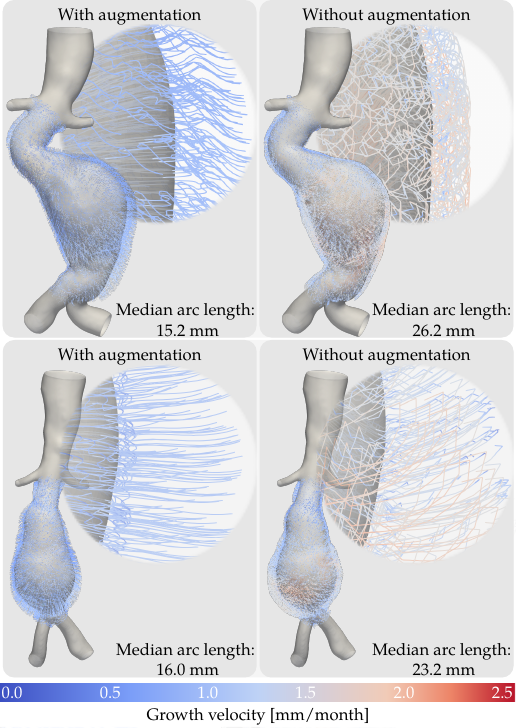}
    \caption{Pointwise AAA growth trajectories over a two-year period, predicted by $g_\theta$ trained with and without temporal augmentation (Section \ref{sec:optimization}). Smoothness of the trajectories is quantified as the median arc length of the trajectories, where smaller is better. The growth prediction model trained with temporal augmentation predicts smoother deformation trajectories than the model trained without temporal augmentation.}
    \label{fig:INR_ablation_qual}
\end{figure}

\subsection{Temporal augmentation ablation}
\begin{figure}[t!]
    \centering
    \includegraphics[width=0.5\textwidth]{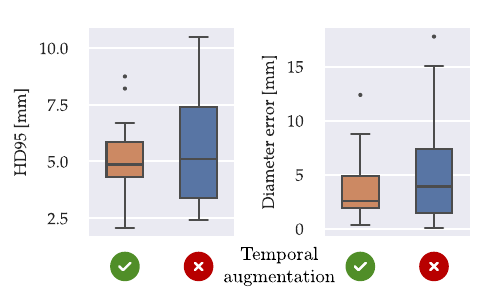}
    \caption{Boxplots containing the HD95 and diameter error for the growth predictions with $\Delta t=$ 24 months using $g_{\theta}$ trained with and without temporal augmentation (Section \ref{sec:optimization}), respectively.}
    \label{fig:INR_ablation_quant}
\end{figure}

We evaluate the added value of temporal augmentation, as introduced in Section \ref{sec:optimization}. We train a growth prediction model without temporal augmentation by sampling two time points from $T_{p_i}$ as reference and target shape, instead of extracting the target shape from $f_{p_i, \theta}$ with a randomly sampled time step. This model is thus trained on a sparse and irregularly spaced set of time steps, as originally present in the longitudinal dataset. We compare the growth predictions between the models trained with and without temporal augmentation. We let both models predict growth over a two-year period with one-month increments, to obtain trajectories for each vertex in the vascular model. We assess the smoothness of the vertex trajectories using the median arc length of the two-year growth trajectories. Figure \ref{fig:INR_ablation_qual} shows the trajectories over two years for two patients and the median arc lengths, where lower is better. Figure \ref{fig:INR_ablation_qual} shows both qualitatively and quantitatively that the prediction model trained with temporal augmentation predicts much smoother trajectories than the model trained without this augmentation. Moreover, the growth velocity, i.e. the vertex deformation in one month, is more constant for the model trained with temporal augmentation. Figure \ref{fig:INR_ablation_quant} shows quantitative performance metrics in terms of the HD95 and diameter error, evaluated for the predictions in two years for each patient. These plots show that temporal augmentation also improves the growth prediction quality.

\begin{table*}[t!]
\caption{Performance metrics for the multi-physical feature ablation, evaluated using HD95, diameter error and RGVD. Medians, IQRs and p-values are shown. The best performing model is highlighted in \textbf{bold}, statistical significant differences (p$<$0.05) compared to the model trained on all features in \textit{italic}. P-values were computed using a Wilcoxon signed-rank test.}
\label{tab:feature_ablation}
\begin{tabular}{ccccccccc}
\hline
\multicolumn{3}{c}{Features} & \multicolumn{2}{c}{HD95 {[}mm{]}} & \multicolumn{2}{c}{Diameter error {[}mm{]}} & \multicolumn{2}{c}{RGVD} \\
$\mathcal{X}^{\text{morph}}$ & $\mathcal{X}^{\text{hemo}}$ & $\mathcal{X}^{\text{hist}}$ & median, IQR & p-value & median, IQR & p-value & median, IQR & p-value \\
\hline
 \checkicon & \checkicon & \checkicon & \textbf{2.75, 1.96} & - & \textbf{1.18, 1.38} & - & 0.10, 0.92 & - \\
\crossicon & \checkicon & \checkicon & 3.16, 1.97 & \textit{0.0061} & 1.43, 3.45 & \textit{0.0045} & -0.22, 1.03 & 0.056 \\
 \checkicon & \crossicon & \checkicon & \textbf{2.75, 1.72} & 0.32 & 1.35, 2.16 & 0.26 & \textbf{-0.08}, 0.94 & 0.87 \\
 \checkicon & \checkicon & \crossicon & 2.88, 1.75 & 0.37 & 1.54, 2.33 & \textit{0.016} & 0.12, 1.30 & 0.61\\
 \hline
\end{tabular}
\end{table*}

\subsection{Multi-physical feature ablation}
We perform an ablation study to evaluate the contribution of each of the multi-physical feature set: morphological ($\mathcal{X}^{\text{morph}}$), hemodynamic ($\mathcal{X}^{\text{hemo}}$) and historical growth features ($\mathcal{X}^{\text{hist}}$) as shown in Figure \ref{fig:method}. We train $g_{\theta}$ on different subsets of these features and evaluate their performance in terms of the HD95, diameter error and RGVD using 5-fold cross validation. We determine the statistical significance of performance differences compared to the model trained on all features using a Wilcoxon signed-rank test. Table \ref{tab:feature_ablation} shows that the inclusion of $\mathcal{X}^{\text{morph}}$ and $\mathcal{X}^{\text{hist}}$ leads to a statistically significant performance gain in two metrics and one metric, respectively. In contrast, including $\mathcal{X}^{\text{hemo}}$ did not result in a statistically significant performance gain for any of the metrics and performs similarly to the model that includes all three feature types.

\begin{figure}[h!]
    \centering
    \begin{subfigure}[t]{0.95\columnwidth}
    \includegraphics[width=\columnwidth]{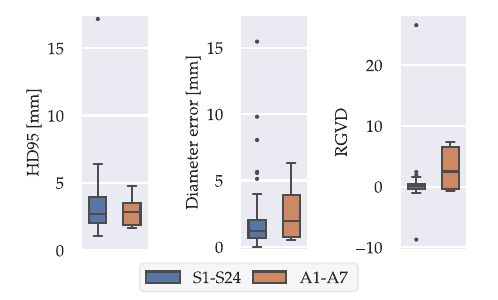}
    \caption{Growth prediction performance for time steps smaller than two years.}
    \label{fig:AMC_quant_a}
    \end{subfigure}
    \begin{subfigure}[t]{0.95\columnwidth}
    \includegraphics[width=\columnwidth]{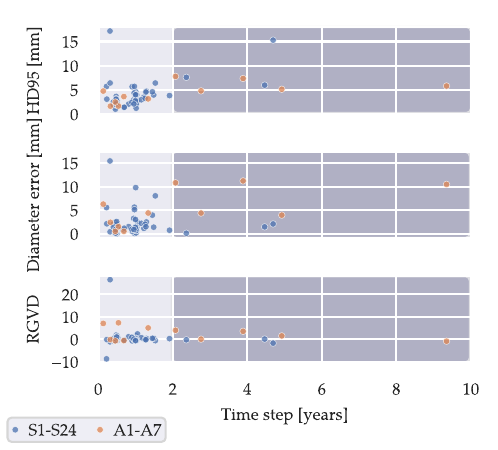}
    \caption{Growth prediction performance as a function of the time steps. The dark gray area indicates time steps that were not observed by the model during training ($> 2$ years).}
    \label{fig:AMC_quant_b}
    \end{subfigure}
    \caption{Growth prediction performance of the ensemble on the external validation set (A1-A7, Figure \ref{fig:data_overview}) and the 5-fold cross validation (S1-S24, Figure \ref{fig:data_overview}) in terms of HD95, diameter error and RGVD.}
\end{figure}

\begin{figure*}[t!]
    \centering
    \includegraphics[scale=1]{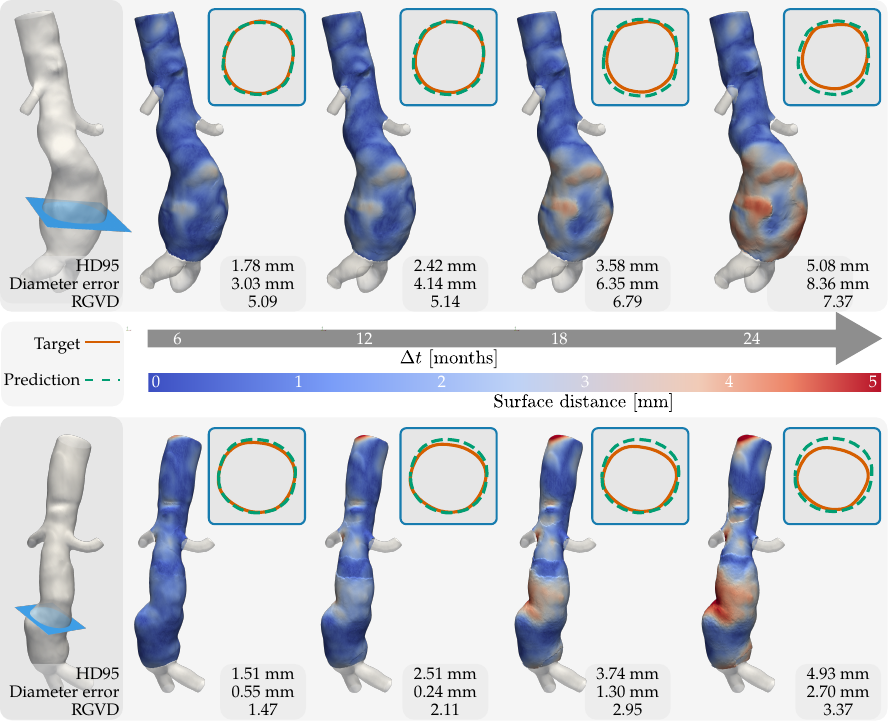}
    \caption{Visualization of growth predictions for two patients in the Amsterdam cohort using standardized time steps ($\Delta t \in \{ 6, 12, 18, 24\}$ months, with local surface indicated on the vascular models. Performance metrics are shown for each growth prediction, as well as 2D cross sections of the target and predicted AAA shape at the level of the indicated blue plane.}
    \label{fig:visual_AMC}
\end{figure*}

\subsection{External validation set}
The previous experiments have evaluated the performance of our growth prediction model $g_{\theta}$ on patients from Seoul National Hospital, South Korea, (S1-S24 Figure \ref{fig:data_overview}) using 5-fold cross validation. To evaluate the generalization of our model to other patient populations, we tested an ensemble of the five trained networks on an external validation set of AAA patients from Amsterdam UMC, The Netherlands (A1-A7, Figure \ref{fig:data_overview}). To aggregate ensemble predictions, we average the orientation and magnitude of the vertex-wise deformations separately.

 We assess the predicted growth between observed time points $t_{j, p_i}$ and $t_{j+1, p_i}$ for $j=1,..,m_{p_i}-2$ for time steps smaller than two years, i.e. time steps observed by the model during training. Figure \ref{fig:AMC_quant_a} presents the ensemble's performance on the vascular models in A1-A7 with time steps below two years ($N=6$) in terms of the HD95, diameter error and RGVD. For reference, we include the results of the cross-validation on S1-S24 for these time steps ($N=62$). The HD95 and diameter errors for A1-A7 and S1-S24 lie in the same range. The RGVD for patients A1-A7 is considerably larger than for S1-S24, implying a tendency of the model to overestimate AAA growth on the external validation cohort.

Our external validation set also contains successive vascular models that were obtained with time steps longer than two years. We analyze the growth prediction performance of our prediction model for these larger time steps as well. Figure \ref{fig:AMC_quant_b} shows the errors as a function of the time step for patients in S1-S24 ($N=65$) and A1-A7 ($N=11$). We observe that for time steps smaller than two years, the HD95 and diameter errors are mostly small and lie within the same range for both datasets. These errors increase in both datasets for time steps beyond this point, implying a limited generalization of the model for larger time steps.

Figure \ref{fig:visual_AMC} shows the qualitative performance of our growth prediction model for fixed time steps \seqsplit{($\Delta t \in \{6, 12, 18, 24\}$)} months for two patients from A1-A7. As observed data is not available at these time points, we have compared the growth predictions to vascular models extracted from $f_{p_i,\theta}$. As expected and in line with the results in Figure \ref{fig:diameter_density}, prediction accuracy declines when the time step size increases. Moreover, as observed in the previous experiments, our model tends to overestimate the AAA growth for patients in this cohort.

\section{Discussion}
We have introduced a geometric deep learning model that predicts personalized growth of AAAs over arbitrary time steps directly on the vascular surface model based on embedded local multi-physical features. We have measured the growth prediction quality in terms of various complementary metrics and showed that our method outperforms two baseline models. Moreover, we have found that our model identifies patients whose AAA will exceed the clinical diameter threshold within two years and are thus eligible for surgery.
Our growth prediction model has the potential to impact healthcare by informing surgeons to identify the right moment for elective surgery, and providing detailed information and visualization to inform patients about their condition and for improved peace of mind \cite{lee2017experience}.

We used a geometric deep learning model to predict AAA growth as 3D deformation vectors acting on the vascular surface. Unlike CNN-based methods that require projecting the vascular surface to a Euclidean grid, \citep{jiang2020deep,kim2022deep,ferdian2022wssnet,gharleghi2020deep,gharleghi2022transient}, our LaB-GATr architecture processes the irregular, non-Euclidean structure of vascular models. These vascular models offer more flexibility in defining the region of interest compared to Euclidean parameterizations, making it easier to include other relevant arteries as input to the model. Here, we have included the common iliac and renal arteries. Although we focus on the deformation of the AAA surface, the location and positioning of the renal and iliac arteries may provide additional useful context for our model. For example, the angles between iliac arteries could influence AAA growth rates \citep{zhaoaortoiliac}.

In this work, we employ LaB-GATr, an SE(3)-symmetric transformer architecture. The preservation of SE(3) symmetry ensures that the model's predictions are invariant to rigid transformations of the vascular geometry in 3D space, improving its data efficiency \cite{suk2023se}. In contrast to geometric deep learning models that leverage graph convolutions or message passing to aggregate information locally across the surface \citep{de2020gauge,sharp2022diffusionnet,suk2024mesh}, our transformer architecture captures global context in a single layer through attention mechanisms. This global approach to information aggregation is more effective on complex geometries such as vascular surfaces \citep{nannini2025learning}.

One surprising result of our experiments is that incorporating hemodynamic features did not significantly improve the performance of our growth prediction model, although a correlation between hemodynamics and AAA growth has been repeatedly suggested \citep{kim2022deep,stevens2017biomechanical}. Previous works have shown the feasibility of using geometric deep learning models to estimate local hemodynamics from vascular models \citep{suk2023se,suk2024physics,suk2024deep}. This suggests a strong connection between the vascular model geometry and hemodynamic quantities on its surface. Hence, explicitly adding hemodynamic features in addition to geometrical features might have limited added value. Notably, the hemodynamic features used in our current model were obtained using standard boundary conditions, and personalized boundary conditions may reveal more information about the subtle influence of hemodynamics on AAA growth. However, to the best of our knowledge, no retrospective longitudinal dataset with personalized boundary condition measurements is available. 

% features
Our model included local multi-physical features that could be derived from image data and that have been hypothesized to influence AAA growth, namely thrombus thickness, local radius, TAWSS, OSI, and historic growth patterns. In addition to these image-derived features, patient-level features, such as smoking status, BMI, age, and sex, have been shown to be associated with AAA growth \citep{brady2004abdominal} and may further improve the model's performance. In future work, these features, where available, could be seamlessly integrated into the prediction model.

Conditioning our model on arbitrary time steps allowed us to assess AAA progression over varying time intervals and to extract clinically relevant parameters from the 3D vascular model. Although the prediction accuracy decreased over time (Figure \ref{fig:diameter_density}), the median diameter error stayed below 2.5 mm for predictions until 24 months, which lies in the same range as the reported intra-observer variability in measuring AAA diameters using CTA \citep{singh2003intra}. We have also shown that by conditioning our model on arbitrary time steps, it can be used to predict if and when the AAA diameter exceeds 55 mm within a two-year time window (Figure \ref{fig:intervals_timeline}). This may impact clinical workflows by tailoring surveillance intervals and the timing of surgical intervention. 

We found that training our growth prediction model with temporal augmentation improved the smoothness of the growth trajectories and ensured more stable growth velocities compared to a model trained without temporal augmentation (Figure \ref{fig:INR_ablation_qual}). Moreover, temporal augmentation led to improved growth prediction quality (Figure \ref{fig:INR_ablation_quant}). Given its ability to model temporally continuous surface deformations from sparse and irregularly sampled data, our model may also be applicable to other pathologies characterized by progressive shape change, such as tumor growth \citep{chen2024vestibular}.

AAA growth is a subtle and complex process, making reliable ground truth data essential for training and evaluating predictive models. However, acquiring the vascular models and multi-physical features from image data requires several pre-processing steps, each of which is a potential source of error. For instance, while previous work has registered vascular models across time points using lumbar vertebrae as anatomical landmarks, \citep{kim2022deep}, we found that aligning the aortic centerlines yielded lower registration errors. Additionally, the irregular time steps in our data introduce additional variation. Although our continuous models $f_{p_i,\theta}$ are designed to handle irregular time steps, longer intervals may increase the uncertainty of interpolated reference geometries. These errors not only affect the growth patterns learned by the model but also the metrics used to evaluate its performance. To mitigate this, we have used several complementary performance metrics that assess different aspects of the growth.

A limitation of our work, which is common to this domain, is the relatively small dataset that was used. In addition to a public training set ($N=24$ patients), we used an external validation set ($N=7$ patients) acquired at a different center. Our results indicated that the method performed differently in both data sets, but the limited data set size prevents us from drawing strong conclusions about this. For example, on average, the AAAs from patients S1-S24 displayed larger growth rates than the AAAs in patients A1-A7, but we cannot statistically confirm that this causes these differences. These datasets were retrospectively collected from hospital databases not specifically curated for AAA growth prediction, resulting in variability in scan intervals and limited control over patient selection. Here, we aimed to address these issues by temporal augmentation and a data-efficient model. However, we anticipate that model performance and generalization would greatly benefit from a carefully designed dataset with a diverse patient population and standardized acquisition protocols, such as in \citep{ristl2022growth}.

The use of geometric deep learning for predicting AAA growth from longitudinal data and multi-physical features opens several promising directions for future research. One option involves leveraging the trained model to simulate hypothetical changes in local geometric or hemodynamic conditions, offering a tool that potentially provides additional insight into individual growth processes. Beyond growth prediction, our framework may be extended to other clinically relevant tasks. For example, prior work has shown that geometric features can be predictive of postoperative complications such as endoleaks \citep{van2024statistical}. Adapting the current model architecture to predict such outcomes could support pre-operative risk assessment and improve clinical decision-making.

\section{Conclusion}
In this work, we have presented a geometric deep learning model that predicts personalized AAA growth patterns over an arbitrary time interval directly on the vascular model, including local, patient-specific features. To handle the irregular temporal spacing inherent in real-world clinical data, we introduced a temporal augmentation strategy that enables the model to learn smooth growth trajectories. Our results demonstrate that it is feasible to estimate localized AAA growth directly on vascular models using geometric deep learning, potentially improving the estimations of individual growth patterns and tailoring of surveillance intervals.

\section{Acknowledgements}
This research was funded by the European Union Horizon Europe Health program \seqsplit{(HORIZON-HLTH-2022-STAYHLTH-01-two-stage)} under the VASCUL-AID project (grant agreement ID: \seqsplit{101080947}).

Jelmer M. Wolterink was supported by the NWO domain Applied and Engineering Sciences VENI grant (18192).

This work made use of the Dutch national e-infrastructure, in particular the Dutch supercomputer Snellius, with the support of the SURF Small Compute Applications grant no. EINF-10199.

We thank the authors of \citep{jiang2020deep,kim2022deep} for kindly sharing their longitudinal AAA dataset with us.


% \section{}\label{}

\printcredits

\bibliographystyle{cas-model2-names}

\bibliography{AAA-refs}



\end{document}